\documentclass[10pt,twocolumn,letterpaper]{article}

\usepackage{cvpr}
\usepackage{times}
\usepackage{epsfig}
\usepackage{graphicx}
\usepackage{amsmath}
\usepackage{amssymb}
\usepackage{algpseudocode}
\usepackage{algorithm}
\usepackage{algcompatible}
\usepackage{multicol}
\usepackage{multirow}
\usepackage{makecell}
\usepackage{hyperref}
\usepackage{bbold}
\usepackage{breqn}
\usepackage{tcolorbox}

\begin{document}

\title{\textit{MorCode}: Face Morphing Attack Generation using Generative Codebooks}

\author{
Aravinda Reddy PN$^{1}$ \and
Raghavendra Ramachandra$^{2}$ \and 
Sushma Venkatesh$^{3}$\and 
Krothapalli Sreenivasa Rao$^{2,1}$ \and
Pabitra Mitra$^{3,1}$ \and
Rakesh Krishna$^{1}$  \and
\\
Indian Institute of Technology Kharagpur$^{1}$, India \\  
Norwegian University of Science and Technology$^{2}$, Norway \\
AiBA AS, Norway$^{3}$
}


\maketitle
\thispagestyle{empty}

\begin{abstract}
Face recognition systems (FRS) can be compromised by face morphing attacks, which blend textural and geometric information from multiple facial images.  The rapid evolution of generative AI, especially Generative Adversarial Networks (GAN) or Diffusion models, where encoded images are interpolated to generate high-quality face morphing images. In this work, we present a novel method for the automatic face morphing generation method \textit{MorCode}, which leverages a contemporary encoder-decoder architecture conditioned on codebook learning to generate high-quality morphing images. Extensive experiments were performed on the newly constructed morphing dataset using five state-of-the-art morphing generation techniques using both digital and print-scan data. The attack potential of the proposed morphing generation technique, \textit{MorCode}, was benchmarked using three different face recognition systems. The obtained results indicate the highest attack potential of the proposed \textit{MorCode}  when compared with five state-of-the-art morphing generation methods on both digital and print scan data.

\end{abstract}

\section{Introduction}
\label{sec:intro}
Face Recognition Systems (FRS) are extensively deployed in various access control applications, including border control, because of their high accuracy and user convenience. Nevertheless, these systems are susceptible to various forms of attacks such as presentation and adversarial attacks, which can compromise their security. Of particular concern are morphing attacks, which have gained prominence for their ability to undermine the security of automatic border-control scenarios. Morphing is the process of blending two or more facial images to result in a single composite facial image that reflects both texture and geometric information corresponding to facial images used for morphing. Therefore, the generated morphing images indicate a vulnerability to human observers, including border guards \cite{godage2022analyzing} and automatic FRS \cite{Venkatesh-MADSurvey-IEEETTS-2021}. NIST Face Analysis Technology Evaluation (FATE) Morph \cite{Nist-Frvt-Morph} illustrates the vulnerability of several FRS to morphing attacks. Therefore, that indicates that, higher the accuracy of FRS implies higher vulnerability.

Face morphing attacks have garnered significant attention, with the objective of maintaining identity-related features from facial images that represent multiple identities. The primary focus is on preserving the individual's identity within the morphed image, as this can enhance the attacker's potential to launch successful morphing attacks on the FRS.  To this extent, researchers have proposed several 2D \cite{Venkatesh-MADSurvey-IEEETTS-2021} and 3D face morphing generation \cite{3DMorphing} algorithms that have indicated a higher attack potential when presented to automatic FRS. The available face morphing generation algorithms can be broadly classified as \cite{Venkatesh-MADSurvey-IEEETTS-2021} (a) facial landmark-based and (b) deep learning-based.  Facial landmark-based methods use pixel information from facial images of multiple identities to obtain the morphing image, whereas deep learning-based methods generate or synthesize the morphing face image based on the compact representation (also called latent) corresponding to multiple identities. Because morphing attacks are applicable to border control scenarios, the goal of morphing generation techniques must be to generate facial images fulfilling the quality constraints laid down by the International Civil Aviation Organization (ICAO), which requires a high-quality facial image.

Early work on generating face morphing images was based on facial landmark-based face morphing techniques. Given two facial images, facial landmark-based methods first extract the facial landmarks (approximately 64 points) after alignment. Triangulation is then performed, followed by wrapping and blending to generate the morphing image. Landmark-based methods include open-source software from open CV \cite{Landmark-face-morph} and FaceMorpher \cite{Facemorpher} which are employed in the literature to generate morphing attacks. One of the major limitations of facial landmark-based methods is ghost artifacts that are prominent in the eye, mouth, and nose regions, hindering image quality issues. Therefore, post-processing using facial landmark-based methods \cite{Landmark-face-morph} further improves the  quality of the morphed images. 

The evolution of generative AI techniques has enabled the generation of face-morphing images that overcome ghost artifacts. Early works used the vanilla GAN \cite{Damer-MorGAN-BTAS-2018}, in which the latent from the facial images to be morphed is averaged to obtain a single latent, which is then used to construct the morphed image. However, the quality of the image rendered in \cite{Damer-MorGAN-BTAS-2018} was low because of the output dimensions of $64 \times 64$ pixels. The first work on generating an ICAO-quality morphing image using StyleGAN was presented in \cite{venkatesh-StyleGAN-morph-IWBF-2020}, which also used latent averaging. However, the use of GAN degrades the identity information, which degrades the attack potential of the generated face-morphing image. To overcome this, MIPGAN \cite{zhang-MIPGAN-TBIOM-2021} was introduced, in which the fusion of the latent from StyleGAN was optimized to achieve the highest attack potential using the FRS as a loss function. MIPGAN has higher attack potential than other GAN-based approaches \cite{zhang-MIPGAN-TBIOM-2021}. The pixel2style2pixel (pSp) encoder-based latent extraction and fusion using spherical interpolation was proposed in \cite{CVMIMorphGeneration}. Finally, the fused latent was used to generate the face morphing image by employing StyleGAN2. The use of the pSp encoder, spherical interpolation, and StyleGAN2 combination indicated a higher attack potential for the morphed image, and it is worth noting that there is no need for identity-based optimization.

The introduction of diffusion models for image generation has attracted researchers to adapt the model for face morphing generation. In \cite{MoDiff} \cite{MoDiff2}, denoising diffusion probabilistic models (DDPM)  is adapted to generate face morphing images by manipulating the latent code. However, identity information is not sufficient to generate high attack potential. Therefore, in \cite{PIPE}, identity-prior-based optimization was introduced to DDPMS to generate morphed images with a higher attack potential. However, the results indicate a marginal increase in attack potential. In \cite{zhang2023morphganformer}, a transformer-based GAN model with identity loss functions was used to generate face morphing images. However, the attack potential of morphing images generated using \cite{zhang2023morphganformer} did not indicate a higher attack potential compared to conventional MorDiff \cite{MoDiff} \cite{MoDiff2}.  As per existing research, it is crucial to demonstrate the vulnerability of the Face Recognition System (FRS) by exhibiting a higher attack potential with the generated face-morphing image. In this regard, it is essential to utilize latent representations in generative networks. Consequently, we were inspired to incorporate a Vector-Quantized Generative Adversarial Network (VQ-GAN) \cite{esser2021taming} into our work, which conditions the latent facial image with codebook to achieve high-quality morphing generation. 

In this work, we introduce a novel method for 2D face morphing generation using a codebook learned using VQ-GAN \cite{esser2021taming} which we refer to as \textit{MorCode}. The novelty of the proposed method lies in the introduction of latent conditioning, which results in a discrete and compact representation of the latent. Thus, it is our assertion that the representation of latent using codebooks will enable a high-quality face morphing generation algorithm. Furthermore, the proposed method employs spherical interpolation to   blend the latent corresponding to multiple face images, which contributes to high-quality morphing generation. The main contributions of this work are as follows: 

\begin{itemize} 
\item Proposed a novel 2D face morphing generation using codebook and spherical interpolation to achieve high quality face morphing generation. 
\item Introduced a new dataset MorCode Morphing Dataset (MMD) with the proposed face morphing generation technique using publicly available face dataset FRGC V2. Newly generated dataset is comprised of 160 data subjects resulting in a total of 1277 bona fide and 2526 morphing images.   
\item Extensive experiments are carried on the newly generated dataset and comparison of the  proposed morphing technique is quantitatively benchmarked with four different existing face morphing techniques using Generalized Morphing Attack Potential (G-MAP) vulnerability metric. The attack potential of the proposed an existing morphing techniques are evaluated against three different deep learning based FRS such as ArcFace \cite{Deng-ArcFace-IEEE-CVPR-2019}, MagFace \cite{magface} and AdaFace \cite{AdaFace}. 
\item The proposed method is open sourced to support the reproducibility.\url{https://github.com/Aravinda27/MorCode}
\end{itemize}

The rest of the paper is organised as follows: Section\ref{sec:Pro} discuss the proposed \textit{MorCode} method for 2D face morphing generation method, Section \ref{sec:db} presents the new dataset generated using the proposed morphing techniques, Section \ref{sec:exp} presents the vulnerability evaluation results of the proposed and existing morphing generation methods against three FRS system and Section \ref{sec:conc} draws the conclusion.

\section{\textit{MorCode}: Proposed 2D Face morphing generation}
\label{sec:Pro}
\begin{figure*}[!h]
\centering
\includegraphics[width=0.8\linewidth]{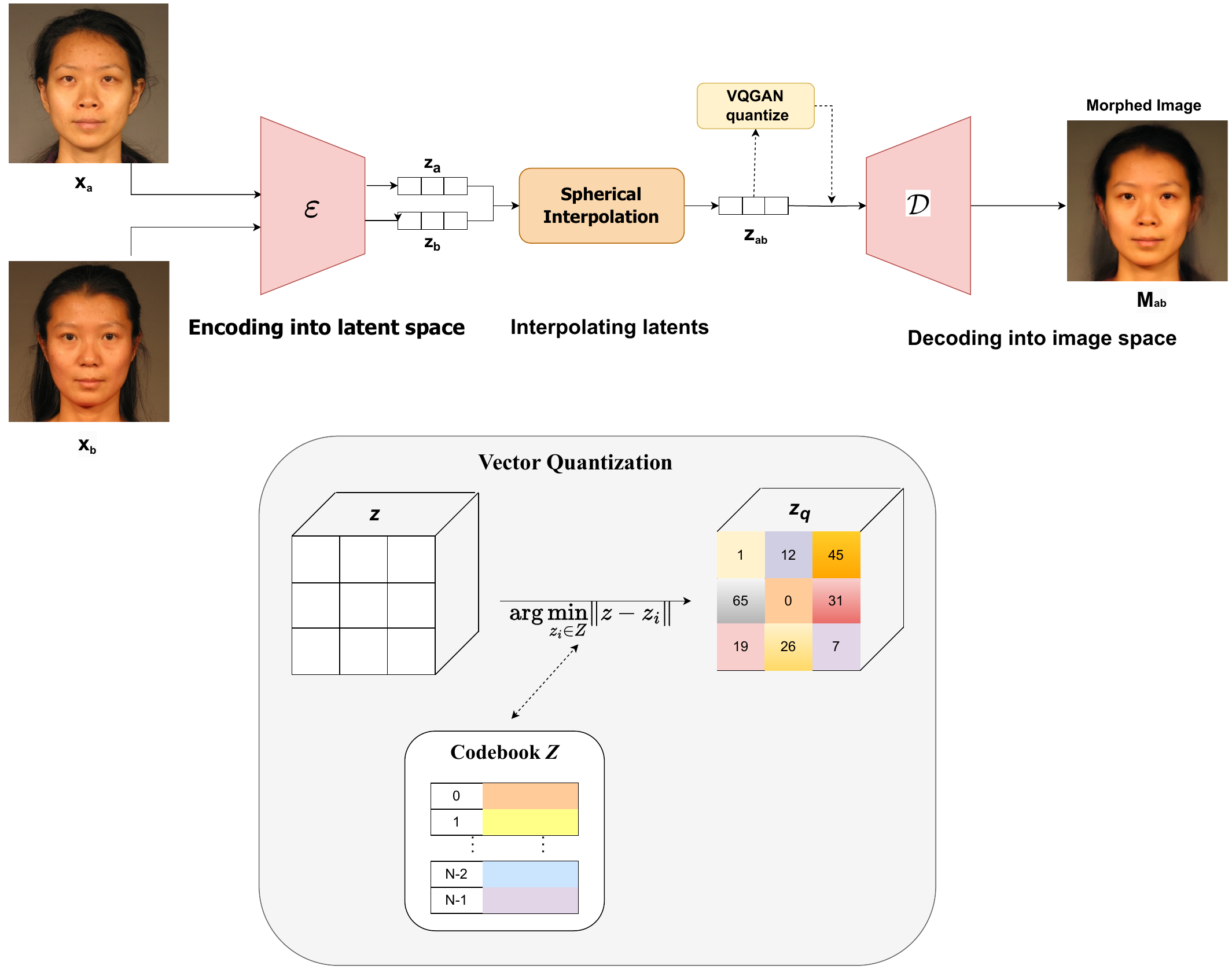}
\caption{Block Diagram of the proposed method \textit{MorCode} to construct high quality morphed images.} \label{fig:block}
\end{figure*}

In this work, we present a new method for generating a high quality 2D face morphing attack using Vector Quantized Generative Adverserial Network (VQGAN). VQGAN \cite{esser2021taming} utilizes a noise-conditioned score network (NCSN++)-based encoder-decoder architecture, which is a score-based generative model. The application of NCSN++ differs from facial morphing; however, its U-Net structure and discrete codebook of acquired representations for each image possess the capability to capture the rich features that are essential for the face morphing generation task. Figure \ref{fig:block} shows the block diagram of the proposed morphing generation technique. The proposed method can be structured in three steps (a) Encoder (b) Spherical Interpolation and (c) Decoder. The underlining idea of the proposed method includes the step from image to latent space manipulation and final image image decoding to ensure the generated morphed image that can preserve the identity information resulting in the higher attack potential. More particularly, the use of perceptually rich codebook in VQGAN allows the representation of the given face image as the spatial collection allows the high quality representation of the latent that enable the high quality image generation \cite{esser2021taming}.

In the following, we present in detail different building blocks of the proposed face morphing generation method. 
\subsection{Encoder}
The encoder architecture consists of serial connection of convolution layer followed by ResNet block and the downsample layer. The use of ResNet blocks will learn the feature representation while preserving spatial dimensions.  Subsequently, a downsampling layer reduces the spatial dimensions of the feature maps, conserving only the most salient features. The downsampling is performed twice, each time followed by two residual blocks to further refine the features. After another single residual block, an attention block (Attn Block) is applied to capture long-range dependencies within the data. The output of the attention block passes through the final residual block before a convolutional layer with three filters (Conv(3)), producing a pre-quantization convolutional representation preQuantConv.  

Given the two facial images $x_a$ and  $x_b$, the encoder will map into the latent space resulting in the latent representation $z_a$ and  $z_b$ respectively with dimensions compressed by the downsampling factor $f$.The encoder facilitates this dimensionality reduction, mapping the high-resolution input images from their original space $\mathbb{R}^{H \times W \times 3}$ to a more manageable latent space $\mathbb{R}^{h \times w \times c}$, where $h = H/f $, $ w = W/f$, and $c$ represents the depth of the latent space. We use $f=4$ and downsample the image dimensions by a factor of $4$ while representing latent space dimensions. 

\begin{equation}
    z_a = E(x_a), \quad z_b = E(x_b)
\end{equation}

\subsection{Spherical Interpolation}
In the next step, we perform spherical interpolation on the latent representations \( z_a \) and \( z_b \). This is a critical step where we generate a single, morphed latent representation \( z_{ab} \) that encapsulates the characteristics of both \( z_a \) and \( z_b \) while ensuring a smooth transition within the latent space.

\begin{equation}
    z_{ab} = slerp(z_a, z_b; \gamma)
\end{equation}

The spherical interpolation utilizes the geodesic path on the unit hypersphere and is mathematically represented as:

\begin{equation}
    z_{ab} = \frac{\sin((1-\gamma)\Omega)}{\sin(\Omega)}z_a + \frac{\sin(\gamma\Omega)}{\sin(\Omega)}z_b
\end{equation}

where \( \Omega \) is the angle between \( z_a \) and \( z_b \), and \( \gamma \) is the interpolation factor that dictates the blend between the two latent representations. The resultant \( z_{ab} \) possesses dimensions analogous to \( z_a \) and \( z_b \), enabling us to handle it within the latent space as we would with any individual latent vector.
\subsection{Vector Quantization}
The spherically interpolated output $ z_{ab}$ is then quantized using a codebook $Z_{k}$. The quantization process maps the continuous latent features to a discrete set of codes within the codebook to obtain the quantized latent representation $\hat z_{ab}$. This quantization allows for the generation of a compact, discrete representation of the interpolated or blended data, which is beneficial for high perceptual image quality reconstruction.
\begin{equation}
    \hat z_{ab} = VQ(z_{ab}) 
\end{equation}

\subsection{Decoder}

The decoder mirrors the encoder's structure but operates in reverse, aiming to reconstruct the input from its quantized representation $ \hat z_{ab}$. Starting with the quantized tensor $ \hat z_{ab}$, the decoder applies a convolutional layer with 512 filters (Conv(512)) and proceeds through a series of upsampling and residual blocks (ResBlock x3, UpSample, ResBlock x3, UpSample, ResBlock x3). Each upsampling step increases the spatial dimensions, enabling the reconstruction of the original image size. An attention block is then used similarly to the encoder, followed by a single residual block and a final convolutional layer with three filters (Conv(3)).  The reconstructed morphed image $M_{ab}$ can be defined as: 
\begin{equation}
    {M}_{ab} = D(\hat z_{ab})
\end{equation}


Figure \ref{fig:DB1} shows an example of a face-morphing image generated using the proposed \textit{MorCode} method. The qualitative results of the proposed method were also compared with  five state-of-the-art face morphing generation methods. The qualitative results indicate the  superior perceptual quality of the proposed \textit{MorCode}, particularly in preserving the shape and texture features. 

\section{MorCode Morphing Dataset (MMD)}
\label{sec:db}

\begin{figure*}
    \centering
    \includegraphics[width=1\linewidth]{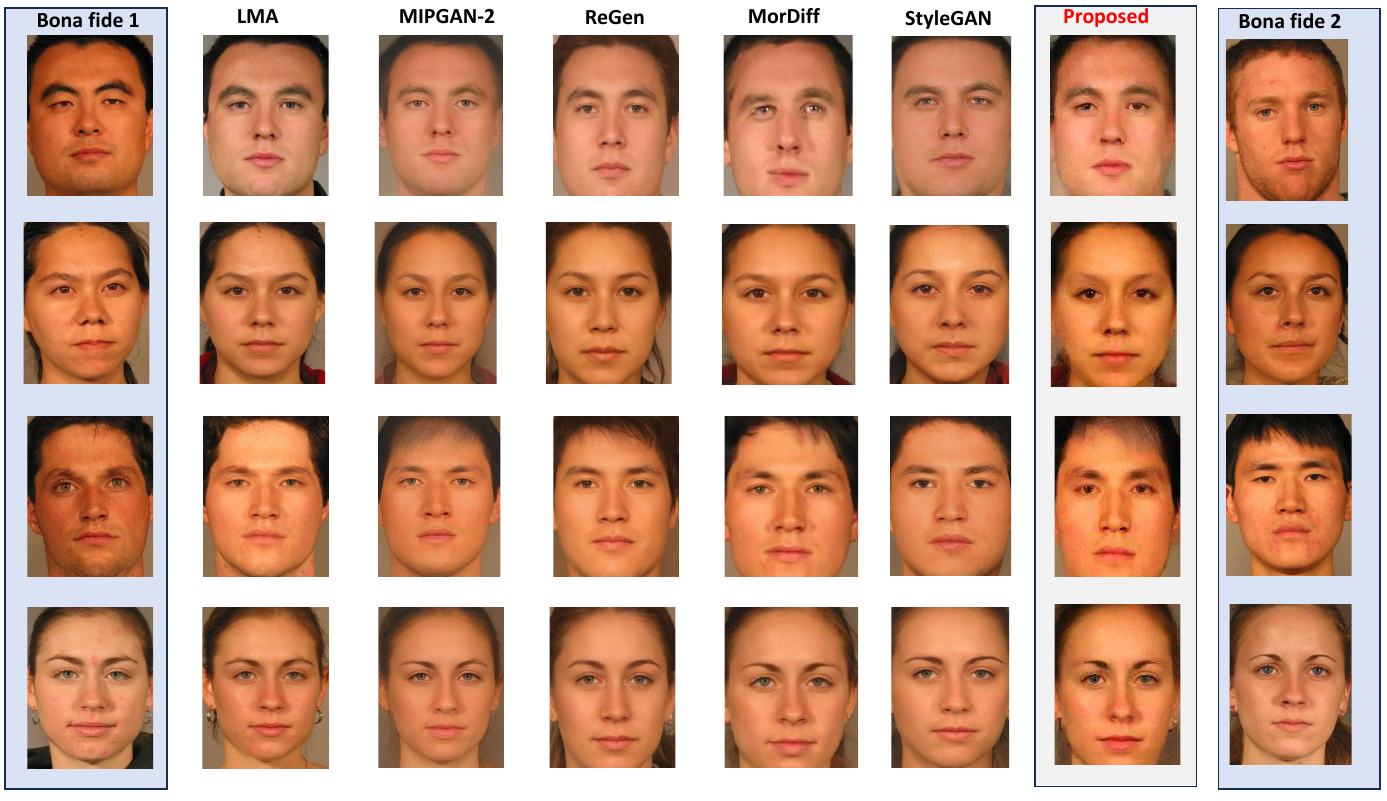}
    \caption{Example images from MMD dataset representing Digital samples. The proposed MorCode face morphing technique is  qualitatively compared with the five different existing techniques.}
    \label{fig:DB1}
\end{figure*}

\begin{figure*}
    \centering
    \includegraphics[width=1\linewidth]{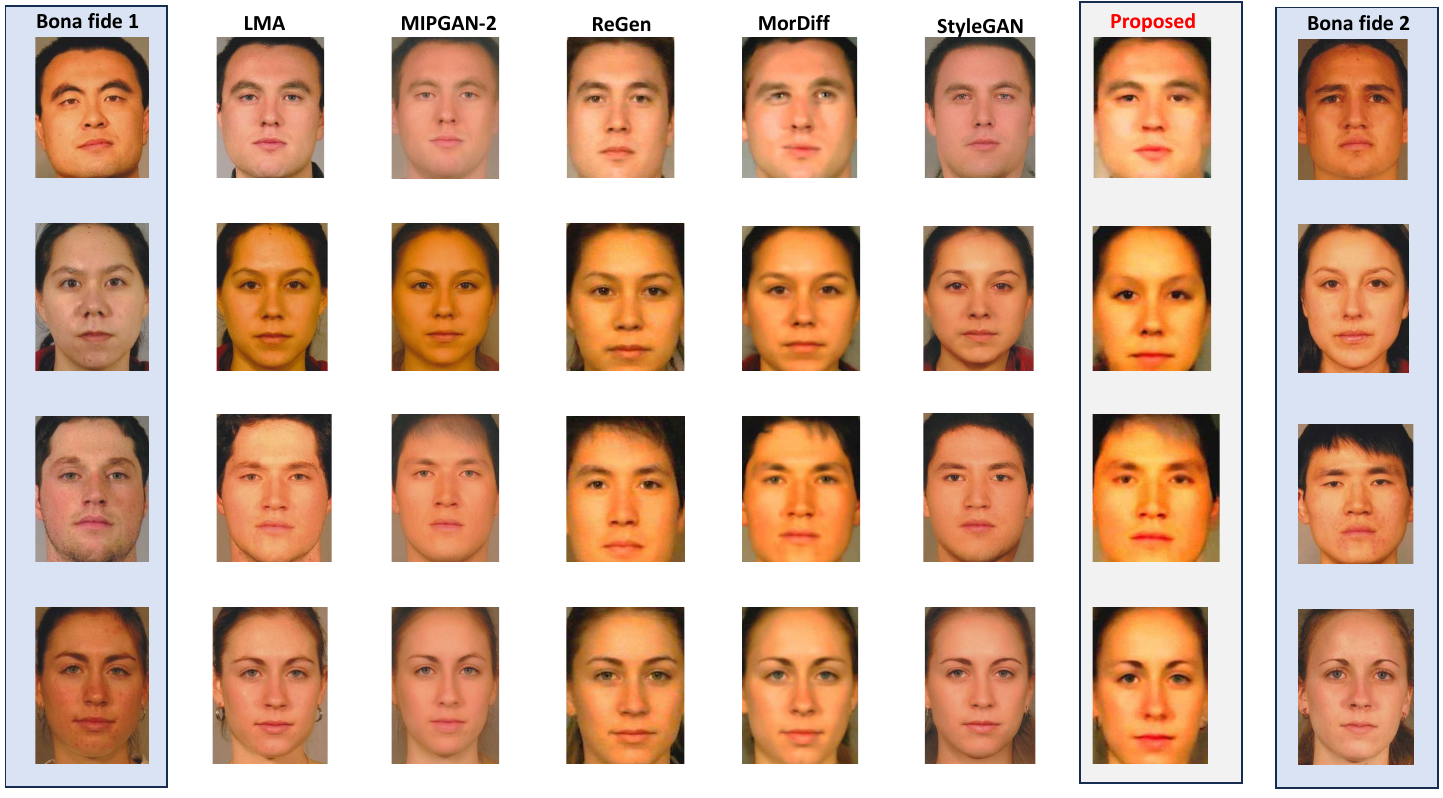}
    \caption{Example images from MMD dataset representing print scan using DNP printer samples. The proposed MorCode face morphing technique is  qualitatively compared with the five different existing techniques.}
    \label{fig:DB2}
\end{figure*}

In this section, we present the MorCode Morphing Dataset (MMD), which is a newly constructed face morphing dataset that utilizes the proposed \textit{MorCode} face morphing generation technique.  MMD dataset was constructed using the publicly available dataset FRGC  V2 \cite{Phillips-OverviewFaceRecognitionGrandChallengeFRGC-CVPR-2005}. We selected 143 subjects with neutral expressions and postures, which were captured under optimal lighting conditions to reflect the conditions of passport enrolment. We followed the recommended procedures for creating face morphs, as outlined in \cite{Raghavendra-FaceMorphingVersusFaceAveraging-IJCB-2017}. To simulate a real-life scenario, we utilized a commercial off-the-shelf product from Neurotek \cite{Neurotechnology-verilook-SDK} to pair face identities based on their closest match. We performed face morphing generation using five existing morphing techniques: landmark-based \cite{Ferrara-TextureBlendingAndShapeWarpingInFaceMorphing-IEEE-BIOSIG-2019}, MIPGAN-2 \cite{zhang-MIPGAN-TBIOM-2021}, ReGen Morph \cite{Damer-ReGenMorphVR-ISVC-2021}, StyleGAN2 \cite{venkatesh-StyleGAN-morph-IWBF-2020} and MorDiff \cite{MoDiff}.

The MMD dataset comprises two types of media: digital and Print-Scan (PS). The digital version encompasses conventional morphing images, whereas the PS morphing images are re-digitized versions of digital morphing images. The inclusion of the PS version was motivated to reflect the passport issuance scenario in which printed passport images were accepted. In this work, we utilized DNP printers, which are specifically designed to produce high-quality passport images featuring facial biometrics. Figure \ref{fig:DB1} and \ref{fig:DB2} show examples from the MMD dataset corresponding to both the Digital and PS datasets. It is worth noting that the quality of the images was slightly degraded by PS. The MMD dataset contains 1276 bona fide samples (separately for digital and morphing) and 2526 morphing images (separately for six different morphing techniques, including the proposed method, and separately for digital and PS).  Therefore, the MMD dataset has $1276 \times 2 = 2552$ bona fide samples and $2526 \times 6 \times 2 = 30312$ morphing images.

\section{Experiments and Results}\label{sec:exp}

In this section, we present a quantitative analysis of the vulnerability of face recognition systems to the proposed \textit{MorCode} morphing generation. The quantitative analysis of the vulnerability was benchmarked using three different deep learning-based FRS that are openly available: ArcFace \cite{Deng-ArcFace-IEEE-CVPR-2019}, MagFace \cite{magface} and AdaFace \cite{AdaFace}. These FRS were selected based on their outstanding verification performance reported in the literature \footnote{We employed Commercial Off-the-Shelf technology (Neurotek \cite{Neurotechnology-verilook-SDK}) to select face image pairs for morphing. To prevent bias in our vulnerability study, we have decided not to use the same COTS. Due to licensing issues involving costs, we are unable to access other Commercial Off-The-Shelf (COTS) FRS. Consequently, we were unable to include the study of COTS within the scope of this research.}. We also compared the performance of the proposed MorCode method with five different state-of-the-art morphing generations: landmark-based \cite{Ferrara-TextureBlendingAndShapeWarpingInFaceMorphing-IEEE-BIOSIG-2019}, MIPGAN-2 \cite{zhang-MIPGAN-TBIOM-2021}, ReGen Morph \cite{Damer-ReGenMorphVR-ISVC-2021}, StyleGAN2 \cite{venkatesh-StyleGAN-morph-IWBF-2020} and MorDiff \cite{MoDiff}.

\begin{table*}[!h]
\centering
\caption{Quantitative performance with G-MAP MA on MMD dataset (Digital).  Higher the value of G-MAP MA indicates the higher vulnerability of FRS to morphing attacks.
}
\label{tab:MADigital}
\resizebox{0.95\columnwidth}{!}{%
\begin{tabular}{|c|c|cc|}
\hline
 &
   &
  \multicolumn{2}{c|}{\textbf{G-MAP with MA}} \\ \cline{3-4} 
 &
   &
  \multicolumn{2}{c|}{\textbf{Operating Threshold: FAR =}} \\ \cline{3-4} 
\multirow{-3}{*}{\textbf{FRS Systems}} &
  \multirow{-3}{*}{\textbf{Morphing Generation Techniques}} &
  \multicolumn{1}{c|}{\textbf{1\%}} &
  \textbf{0.1\%} \\ \hline
 &
  landmark-based   \cite{Ferrara-TextureBlendingAndShapeWarpingInFaceMorphing-IEEE-BIOSIG-2019} &
  \multicolumn{1}{c|}{84.67} &
  17.29 \\ \cline{2-4} 
 &
  MIPGAN-2   \cite{zhang-MIPGAN-TBIOM-2021} &
  \multicolumn{1}{c|}{91.45} &
  15.23 \\ \cline{2-4} 
 &
  ReGen Morph   \cite{Damer-ReGenMorphVR-ISVC-2021} &
  \multicolumn{1}{c|}{29.92} &
  0 \\ \cline{2-4} 
 &
  MorDiff \cite{MoDiff} &
  \multicolumn{1}{c|}{90.81} &
  17.47 \\ \cline{2-4} 
 &
  StyleGAN2   \cite{venkatesh-StyleGAN-morph-IWBF-2020} &
  \multicolumn{1}{c|}{67.99} &
  1.89 \\ \cline{2-4} 
\multirow{-6}{*}{AdaFace \cite{AdaFace}} &
  \textbf{Proposed   method (MorCode)} &
  \multicolumn{1}{c|}{{\color[HTML]{3166FF} \textbf{93.22}}} &
  {\color[HTML]{3166FF} \textbf{19.62}} \\ \hline \hline
 &
  landmark-based   \cite{Ferrara-TextureBlendingAndShapeWarpingInFaceMorphing-IEEE-BIOSIG-2019} &
  \multicolumn{1}{c|}{85.63} &
  26.20 \\ \cline{2-4} 
 &
  MIPGAN-2   \cite{zhang-MIPGAN-TBIOM-2021} &
  \multicolumn{1}{c|}{90.12} &
  24.59 \\ \cline{2-4} 
 &
  ReGen Morph   \cite{Damer-ReGenMorphVR-ISVC-2021} &
  \multicolumn{1}{c|}{33.44} &
  0.04 \\ \cline{2-4} 
 &
  MorDiff \cite{MoDiff} &
  \multicolumn{1}{c|}{84.09} &
  15.74 \\ \cline{2-4} 
 &
  StyleGAN2   \cite{venkatesh-StyleGAN-morph-IWBF-2020} &
  \multicolumn{1}{c|}{73.30} &
  3.78 \\ \cline{2-4} 
\multirow{-6}{*}{ArcFace \cite{Deng-ArcFace-IEEE-CVPR-2019}} &
  \textbf{Proposed   method (MorCode)} &
  \multicolumn{1}{c|}{{\color[HTML]{3166FF} \textbf{92.25}}} &
  {\color[HTML]{3166FF} \textbf{27.52}} \\ \hline\hline
 &
  landmark-based   \cite{Ferrara-TextureBlendingAndShapeWarpingInFaceMorphing-IEEE-BIOSIG-2019} &
  \multicolumn{1}{c|}{92.21} &
  38.49 \\ \cline{2-4} 
 &
  MIPGAN-2   \cite{zhang-MIPGAN-TBIOM-2021} &
  \multicolumn{1}{c|}{93.56} &
  36.44 \\ \cline{2-4} 
 &
  ReGen Morph   \cite{Damer-ReGenMorphVR-ISVC-2021} &
  \multicolumn{1}{c|}{64.19} &
  0.17 \\ \cline{2-4} 
 &
  MorDiff \cite{MoDiff} &
  \multicolumn{1}{c|}{93.18} &
  33.57 \\ \cline{2-4} 
 &
  StyleGAN2   \cite{venkatesh-StyleGAN-morph-IWBF-2020} &
  \multicolumn{1}{c|}{83.79} &
  8.58 \\ \cline{2-4} 
\multirow{-6}{*}{MagFace \cite{magface}} &
  \textbf{Proposed   method (MorCode)} &
  \multicolumn{1}{c|}{{\color[HTML]{3166FF} \textbf{96.27}}} &
  {\color[HTML]{3166FF} \textbf{39.48}} \\ \hline
\end{tabular}%
}
\end{table*}
\begin{table*}[!h]
\centering
\caption{{Quantitative performance with G-MAP MA on PS version of MMD dataset.  Higher the value of G-MAP MA indicates the higher vulnerability of FRS to morphing attacks.
}}
\label{tab:MAPrint}
\resizebox{0.95\columnwidth}{!}{%
\begin{tabular}{|c|c|cc|}
\hline
 &
   &
  \multicolumn{2}{c|}{\textbf{G-MAP with MA}} \\ \cline{3-4} 
 &
   &
  \multicolumn{2}{c|}{\textbf{Operating Threshold: FAR =}} \\ \cline{3-4} 
\multirow{-3}{*}{\textbf{FRS Systems}} &
  \multirow{-3}{*}{\textbf{Morphing Generation Techniques}} &
  \multicolumn{1}{c|}{1\%} &
  0.1\% \\ \hline
 &
  landmark-based   \cite{Ferrara-TextureBlendingAndShapeWarpingInFaceMorphing-IEEE-BIOSIG-2019} &
  \multicolumn{1}{c|}{88.41} &
  {\color[HTML]{3166FF} \textbf{17.34}} \\ \cline{2-4} 
 &
  MIPGAN-2   \cite{zhang-MIPGAN-TBIOM-2021} &
  \multicolumn{1}{c|}{92.16} &
  13.85 \\ \cline{2-4} 
 &
  ReGen Morph   \cite{Damer-ReGenMorphVR-ISVC-2021} &
  \multicolumn{1}{c|}{35.32} &
  0.24 \\ \cline{2-4} 
 &
  MorDiff \cite{MoDiff} &
  \multicolumn{1}{c|}{92.40} &
  21.57 \\ \cline{2-4} 
 &
  StyleGAN2   \cite{venkatesh-StyleGAN-morph-IWBF-2020} &
  \multicolumn{1}{c|}{73.61} &
  3.22 \\ \cline{2-4} 
\multirow{-6}{*}{AdaFace \cite{AdaFace}} &
  \textbf{Proposed   method (MorCode)} &
  \multicolumn{1}{c|}{{\color[HTML]{3166FF} \textbf{93.26}}} &
  6.61 \\ \hline \hline
 &
  landmark-based   \cite{Ferrara-TextureBlendingAndShapeWarpingInFaceMorphing-IEEE-BIOSIG-2019} &
  \multicolumn{1}{c|}{90.14} &
  {\color[HTML]{3166FF} \textbf{25.34}} \\ \cline{2-4} 
 &
  MIPGAN-2   \cite{zhang-MIPGAN-TBIOM-2021} &
  \multicolumn{1}{c|}{92.57} &
  22.67 \\ \cline{2-4} 
 &
  ReGen Morph   \cite{Damer-ReGenMorphVR-ISVC-2021} &
  \multicolumn{1}{c|}{42.93} &
  0.55 \\ \cline{2-4} 
 &
  MorDiff \cite{MoDiff} &
  \multicolumn{1}{c|}{88.15} &
  23.15 \\ \cline{2-4} 
 &
  StyleGAN2   \cite{venkatesh-StyleGAN-morph-IWBF-2020} &
  \multicolumn{1}{c|}{75.99} &
  5.54 \\ \cline{2-4} 
\multirow{-6}{*}{ArcFace \cite{Deng-ArcFace-IEEE-CVPR-2019}} &
  \textbf{Proposed   method (MorCode)} &
  \multicolumn{1}{c|}{{\color[HTML]{3166FF} \textbf{94.15}}} &
  5.49 \\ \hline \hline
 &
  landmark-based   \cite{Ferrara-TextureBlendingAndShapeWarpingInFaceMorphing-IEEE-BIOSIG-2019} &
  \multicolumn{1}{c|}{91.14} &
  {\color[HTML]{3166FF} \textbf{39.16}} \\ \cline{2-4} 
 &
  MIPGAN-2   \cite{zhang-MIPGAN-TBIOM-2021} &
  \multicolumn{1}{c|}{92.58} &
  34.71 \\ \cline{2-4} 
 &
  ReGen Morph   \cite{Damer-ReGenMorphVR-ISVC-2021} &
  \multicolumn{1}{c|}{58.98} &
  1.21 \\ \cline{2-4} 
 &
  MorDiff \cite{MoDiff} &
  \multicolumn{1}{c|}{94.17} &
  33.34 \\ \cline{2-4} 
 &
  StyleGAN2   \cite{venkatesh-StyleGAN-morph-IWBF-2020} &
  \multicolumn{1}{c|}{86.69} &
  13.13 \\ \cline{2-4} 
\multirow{-6}{*}{MagFace \cite{magface}} &
  \textbf{Proposed   method (MorCode)} &
  \multicolumn{1}{c|}{{\color[HTML]{3166FF} \textbf{93.22}}} &
  21.13 \\ \hline
\end{tabular}%
}
\end{table*}
There are four different metrics that are employed to benchmark the vulnerability (a) Mated Morph Presentation Match Rate (MMPMR)\cite{Scherhag-MorphingAttacks-MorphingTechniques-BIOSIG-2017} (b) Fully Mated Morph Presentation Match Rate (FMMPMR)\cite{venkatesh-newborn-morphing-IJCB-2021} (c) Morphing Attack Potential (MAP) \cite{ISOMap2023}   and (d) Generalized Morphing Attack Potential
(G-MAP)\cite{GMAP}.
In this work, we quantify the attack potential of morphing generation methods using the Generalized Morphing Attack Potential (G-MAP) vulnerability metric, as it is designed to address the limitations of other evaluation metrics, as discussed in \cite{GMAP} .\footnote{Readers can refer \cite{GMAP} and Table  4 (in \cite{GMAP}) to embrace the  more information on G-MAP.} The G-MAP can be computed as follows\cite{GMAP} \footnote{Taken from \cite{GMAP}}: 
\begin{equation}\label{eqn:G-MAP}
\begin{aligned}
&{\textrm{G-MAP}} ={\frac{1}{|\mathbb{D}|}}{\sum_{d}^{|\mathbb{D}|}}{\frac{1}{|\mathbb{P}|}}{\frac{1}{|\mathbb{M}_d|}} \min_{{\mathbb{F}_l}}\\
& \sum_{i,j}^{|\mathbb{P}|,|\mathbb{M}_d|}\bigg\{\left[ (S1_i^j > \tau_l) \wedge \cdots ( Sk_i^j > \tau_l) \right]\\ &{\times}  \left[ (1-FTAR(i,l)) \right] \bigg\}\\
\end{aligned}
\end{equation}
Where, $\mathbb{P}$ denote the set of paired probe images, $\mathbb{F}$ denote the set of FRS, $\mathbb{D}$ denote the set of Morphing Attack Generation Type, $\mathbb{M}_d$ denote the face morphing image set corresponding to Morphing Attack Generation Type $d$, $\tau_l$ indicate the similarity score threshold for FRS ($l$),$||$ represents the count of elements in a set during metric evaluation and $FTAR(i,l)$ is the failure to acquire probe image in attempt $i$ using  FRS ($l$). 
In this work, we present two results by varying the parameters of G-MAP, as mentioned in \cite{GMAP} (a) G-MAP with Multiple probe Attempts (MA) by setting $\mathbb{D}$ and $\mathbb{{F}_l}$  to $1$ (b) G-MAP with MA and multiple FRS (G-MAP MA and MFRS) by setting $\mathbb{D} = 1$. In both experiments, we set $FTAR = 0$.

To compute the vulnerability of the FRS (or attack potential of morphing generation techniques), we enrol the morphing image and probe the identities that are used to generate the morphing image. The probe facial images correspond to different independent attempts made by the individual identity. A morphing image is considered vulnerable if the probe attempts made by all identities exceed the preset threshold  at the given FAR. In this study, we used the preset thresholds of the FRS with FAR = 1\% and 0.1\%.  Therefore, the higher the value of G-MAP, the higher is the vulnerability of the FRS for the given morphing attack generation technique.

Tables \ref{tab:MADigital} and \ref{tab:MAPrint} show the quantitative benchmarks of the vulnerability of three different FRS using G-MAP MA for digital and print-scan morphing data. Based on the obtained results, the following can be observed. 
\begin{itemize}
    \item The vulnerability performance of the FRS varies across digital and PS-morphing images. MagFace \cite{magface} FRS indicates the highest vulnerability in a digital database with a G-MAP MA of $96.27\%$ at FAR = $1\%$. With the PS database, all three FRS indicated a similar performance in which ArcFace \cite{Deng-ArcFace-IEEE-CVPR-2019} indicated a marginally higher vulnerability at FAR = $1\%$. However, with the lower FAR values ($0.1\%$), all FRS indicate lower vulnerability, and among different FRS, MagFace \cite{magface} indicates a higher vulnerability. 

    \item The proposed \textit{MorCode} indicates the highest attack potential compared to five different morphing generation techniques on the digital morphing images. All three FRS indicate the highest vulnerability for the proposed \textit{MorCode} with G-MAP MA = $93.22\%$ (for AdaFace), $92.25\%$ (for ArcFace), and $96.27\%$ (for MagFace), with FAR = $1\%$. Similar observations can also be made with a lower FAR value of $0.1\%$. Among the three different FRS, MagFace \cite{magface} indicates a higher vulnerability for the proposed \textit{MorCode}, irrespective of the operating threshold. The improved attack potential of the proposed \textit{MorCode} can be attributed to the shape and texture information captured from the identities that contribute to morphing generation.

\item The proposed \textit{MorCode} indicates the higher vulnerability on all three FRS especially at higher FAR values. The attack potential of the proposed method at FAR = $1\%$ indicates similar performance across all three FRS. However, at FAR = $0.1\%$, the proposed method indicates degraded performance, which can be attributed to the granular noise in the morphing image being further enhanced during the print-scan operation. The effect of the print-scan process on the proposed methods can also be visualized in Figure \ref{fig:DB2}, where the degradation in the image quality is noted more compared to other existing morphing techniques. 

\end{itemize}

\begin{table}[!h]
\centering
\caption{Quantitative performance with  G-MAP with MA and multiple FRS on MMD dataset (digital). Higher the value of G-MAP MA with MFRS indicates the higher vulnerability of FRS to morphing attacks.}
\label{tab:MFRS1}
\resizebox{0.88\columnwidth}{!}{
\begin{tabular}{|c|cc|}
\hline
                                                                   & \multicolumn{2}{c|}{\textbf{G-MAP with MA and   MFRS}}      \\ \cline{2-3} 
                                                                   & \multicolumn{2}{c|}{\textbf{Operating Threshold:   FAR =}} \\ \cline{2-3} 
\multirow{-3}{*}{\textbf{Morphing Generation   Techniques}} &
  \multicolumn{1}{c|}{\textbf{1\%}} &
  \textbf{0.1\%} \\ \hline
Landmark-based   \cite{Ferrara-TextureBlendingAndShapeWarpingInFaceMorphing-IEEE-BIOSIG-2019}          & \multicolumn{1}{c|}{84.67}              & 17.29           \\ \hline
MIPGAN-2   \cite{zhang-MIPGAN-TBIOM-2021}         & \multicolumn{1}{c|}{90.12}              & 15.23            \\ \hline
ReGen Morph   \cite{Damer-ReGenMorphVR-ISVC-2021} & \multicolumn{1}{c|}{29.92}              & 0                \\ \hline
MorDiff   \cite{MoDiff}                           & \multicolumn{1}{c|}{84.09}              & 15.74           \\ \hline
StyleGAN2   \cite{venkatesh-StyleGAN-morph-IWBF-2020} &
  \multicolumn{1}{c|}{67.99} &
  1.89 \\ \hline
\textbf{Proposed method (MorCode)} &
  \multicolumn{1}{c|}{{\color[HTML]{3166FF} \textbf{92.25}}} &
  {\color[HTML]{3166FF} \textbf{19.52}} \\ \hline
\end{tabular}%
}
\end{table}

\begin{table}[!h]
\centering
\caption{Quantitative performance with  G-MAP with MA and multiple FRS on MMD dataset (PS). Higher the value of G-MAP MA with MFRS indicates the higher vulnerability of FRS to morphing attacks.}
\label{tab:MFRS2}
\resizebox{.88\columnwidth}{!}{%
\begin{tabular}{|c|cc|}
\hline
                                                          & \multicolumn{2}{c|}{\textbf{G-MAP with MA and   MFRS}}              \\ \cline{2-3} 
                                                          & \multicolumn{2}{c|}{\textbf{Operating Threshold:   FAR =}}         \\ \cline{2-3} 
\multirow{-3}{*}{\textbf{Morphing Generation   Techniques}}            & \multicolumn{1}{c|}{\textbf{1\%}} & \textbf{0.1\%}                        \\ \hline
Landmark-based   \cite{Ferrara-TextureBlendingAndShapeWarpingInFaceMorphing-IEEE-BIOSIG-2019} & \multicolumn{1}{c|}{88.41}                                 & 17.34 \\ \hline
MIPGAN-2   \cite{zhang-MIPGAN-TBIOM-2021}             & \multicolumn{1}{c|}{92.16}        & {13.85} \\ \hline
ReGen Morph   \cite{Damer-ReGenMorphVR-ISVC-2021}     & \multicolumn{1}{c|}{35.32}        & {0.24} \\ \hline
MorDiff   \cite{MoDiff}                  & \multicolumn{1}{c|}{88.15}                                 & {\color[HTML]{3166FF} \textbf{21.57}}     \\ \hline
StyleGAN2   \cite{venkatesh-StyleGAN-morph-IWBF-2020} & \multicolumn{1}{c|}{73.61}        & 3.22                                \\ \hline
\textbf{Proposed method (MorCode)}                                 & \multicolumn{1}{c|}{{\color[HTML]{3166FF} \textbf{93.22}}} & 5.49 \\ \hline
\end{tabular}%
}
\end{table}

Tables \ref{tab:MFRS1} and \ref{tab:MFRS2} illustrate the attack potential of morphing generation techniques, which are quantified using G-MAP with MA and multiple FRS. In this work, the given morphing image is considered to have attack potential if it can successfully deceive all three FRS that were used in this work when probed (with various attempts) with the identities that were used to generate the morphing images. Based on the obtained results following can be noted: 
\begin{itemize}
   
   \item The proposed \textit{MorCode} has indicated a highest attack potential on digital MMD dataset compared to the five different state-of-the art morphing generation techniques. The proposed method indicates an attack potential of G-MAP (MA and MFRS)  = $92.25\%$ and $19.52\%$ at FAR = $1\%$ and FAR = $0.1\%$ respectively. The second-best performance was noted with the MIPGAN-2 \cite{zhang-MIPGAN-TBIOM-2021} morphing technique.

\item With PS version of MMD dataset, the proposed \textit{MorCode} indicates the highest attack potential at higher FAR = $1\%$ with G-MAP (MA and MFRS) = $93.22\%$ . However, at lower FAR values, the performance of the proposed method indicated degraded results. MorDiff \cite{MoDiff,MoDiff2} demonstrated the best performance with G-MAP (MA and MFRS) = $27.57\%$ at FAR = $0.1\%$.

\item It is interesting to note that, the proposed method has indicated the similar attack potential across both digital and PS medium at lower FAR = $1\%$.

\item When compared to the landmarks based morphing generation, the generative deep learning based techniques have indicates higher vulnerability as indicated in the Table \ref{tab:MFRS1} and \ref{tab:MFRS2}. 

\end{itemize}

\section{Conclusion}
\label{sec:conc}
In this work, we presented a novel method,  \textit{MorCode}, to generate a face-morphing attack that can be instrumented on a face recognition system. The proposed \textit{MorCode} is designed using a modern encoder-decoder-based architecture conditioned on VQGAN to generate morphing attacks by spherically interpolating the encoded latent of the two face images to be morphed, which are later decoded to generate the final morphing image. Extensive experiments were performed on the newly constructed morphing dataset using digital and print-scan data with five state-of-the-art face morphing generation methods.  The attack potential of the proposed \textit{MorCode} was benchmarked on both digital and print-scan morphing datasets and compared against five different stat-of-morphing generations. The quantitative results indicated the best attack potential when tested against the three different face recognition systems.The outcomes of our study demonstrate the effectiveness of the \textit{MorCode}-driven method for face-morphing generation.


\end{document}


\title{ Gumbel Rao Monte Carlo based Bi-Modal Neural Architecture Search for Audio-Visual Deepfake Detection\\ \hspace{3.5cm} Supplementary material}


\maketitle
\thispagestyle{empty}

\setcounter{section}{0}
\renewcommand{\thesection}{\Alph{section}}

\section{Gumbel-Rao Monte Carlo estimator}

\begin{figure*}[!h]
  \centering
   \includegraphics[width=0.5\linewidth]{images/grmc.png}

   \caption{Gradient estimation in GRMC estimator. Black arrows indicates the forward pass and red arrows indicate the backward path}
   \label{fig:grmc_png}
\end{figure*}

Figure \ref{fig:grmc_png} illustrates the process of gradient estimation using the GRMC estimator. During the forward pass, Gumbel sampling is executed, followed by the application of the Rao-Blackwellisation condition, and finally, an average is taken over the Monte Carlo samples. In contrast, the backward pass utilizes the straight-through estimator.

\section{Datasets}
\subsubsection{FakeAVCeleb:} The FakeAVCeleb  built using real videos from VoxCeleb2. VoxCeleb2 is a collection of YouTube videos featuring 6,112 celebrities. It includes more than 1 million videos in the development set and more than 36,000 in the test set, each video containing celebrity interviews. To ensure diversity, videos were carefully selected 500 videos (one per celebrity) from VoxCeleb2, averaging 7.8 seconds each. They maintain the original video dimensions and aim for balanced representation across gender, ethnicity, and age. The database features individuals from four ethnicities (Caucasian, Black, South Asian, East Asian) with 100 videos each, equally divided between males and females. The deepfake dataset, FakeAVCeleb, was created, starting with 500 real celebrities videos. Using various deepfake techniques like face-swapping and reenactment, they generated roughly 20,000 manipulated videos. From these original 500 videos, fake video samples were generated using four of the most recent and advanced deepfake manipulation and synthetic speech generation techniques available at the time. These techniques include Faceswap, Faceswap GAN(FSGAN), Wav2Lip, and Real-Time-VoiceCloning (RTVC). These techniques were applied in combination to create a diverse set of fake video samples. Consequently, the database comprises a total of 20,000 video samples, encompassing both original videos featuring celebrities and synthetic deepfake videos generated using the mentioned techniques.

\subsubsection{SWAN-DF:} SWAN-DF is the first high fidelity publicly available dataset of realistic audio-visual deepfakes, where both faces and voices appear and sound like the target person. The SWAN-DF database is based on the public SWAN database of real videos recorded in HD with the iPhone and iPad Pro in 2019. The original SWAN database comprises 150 subjects, such that 50 data subjects were collected in Norway, 50 data subjects were collected in Switzerland, 5 data subjects in France, and 45 data subjects in India. The age distribution of the data subjects is between 20 and 60 years. The dataset was collected using Apple iPhone6S and Apple iPad Pro (12.9inch iPad Pro). Using a combination of autoencoder-based face-swapping models and blending strategies from DeepFaceLab, the faces of the individuals in each pair were swapped to create deepfake videos. Additionally, voice conversion techniques, also known as voice cloning, were employed to manipulate the audio of the videos. Techniques such as zero-shot YourTTS, DiffVC, HiFiVC, and multiple models from FreeVC were utilized for this purpose.

In total, the SWAN-DF database comprises 24,000 deepfake videos, with each video generated using different combinations of face-swapping models and voice conversion techniques. These videos are available in three different pixel resolutions: 
$160\times160$, $256\times 256$, and $320\times 320$. Alongside the deepfake videos, the database also includes 2,800 real videos, likely serving as the ground truth or reference for comparison purposes.

\subsection{Different Augumentation techniques}
\begin{itemize}
    \item \textbf{RandomRotate:} Rotate video randomly by a random angle within given bounds.
    \item \textbf{RandomResize:} Resize the video by zooming in and out.
    \item \textbf{RandomTranslate:} Shifting video in x and y coordinates.
    \item \textbf{RandomShear:} Shearing video in X and Y directions.
    \item \textbf{CenterCrop:} Extract the center crop of the video.
    \item \textbf{CornerCrop:} Extract the corner crop of the video.
    \item \textbf{RandomCrop:} Extract random crop of the video.
    \item \textbf{HorizontalFlip:} Horizontally flip the video
    \item \textbf{VerticalFlip:} Vertically flip the video
    \item \textbf{GaussianBlur:} Augmenter to blur images using Gaussian kernels.
    \item \textbf{ElasticTransformation:} Augmenter to transform images by moving pixels locally around using displacement fields.
    \item \textbf{PiecewiseAffineTransform:} Augmenter that places a regular grid of points on an image and randomly moves the neighborhood of these point around via affine transformations.
    \item \textbf{Superpixel:} Completely or partially transform images to their superpixel representation.
    \item \textbf{Sequential:} Composes several augmentations together.
    \item \textbf{Add:} Add a value to all pixel intensities in a video.
    \item \textbf{Multiply:} Multiply all pixel intensities with the given value. This augmenter can be used to make images lighter or darker
    \item \textbf{Pepper:} Augmenter that sets a certain fraction of pixel intensities to 0, hence they become black.
    \item \textbf{Salt:} Augmenter that sets a certain fraction of pixel intensities to 255, hence they become white.
\end{itemize}

  
We have applied these 13 different augmentations with different values to make 18 different augmentations except the horizontal flip. Then we horizontally flipped all the 18 augmented videos to get 36 different real augmented videos.
After applying the above-mentioned augmentation techniques for training and validation videos in the
FakeAvCeleb and SWAN-DF datasets we have the split shown in Table \ref{tab:data_aug} and the final train and validation and test split is given by the Table \ref{tab:total_aug_final}. The sample images obtained after augumentation is shown in Figure \ref{fig:eleven_images}.

\subsection{Pre-processing}

\begin{itemize}
    \item \textbf{Video pre-processing:}
    \begin{itemize}
        \item After splitting the database into FakeAvCeleb and SWAN-DF real videos, 36 different augmentations were applied to the real videos in the FakeAvCeleb database, while the SWAN-DF real videos underwent 6 augmentations.
        \item The augmentation applied are salt, pepper, shear, changing intensity, horizontal flip, and many more to balance the database.
        \item Converted all videos of the database to $256\times256$ resolution images with 30fps.
        \item To perform face detection, MTCNN is utilised. MTCNN utilizes
a cascaded framework for joint face detection and alignment.
\item A pre-trained ResNet-34 is utilised to extract the facial image features
    \end{itemize}
    \item \textbf{Audio pre-processing:}
    \begin{itemize}
        \item We extract the audio from video files and and subsequently converting the sampling rate of the audio files to 16,000Hz.
        \item We normalize the values of the audio file by applying mean and convert it to mel spectrogram with window length of 25ms and a shift size of 10ms.
        \item In order to standardize the size of all spectrograms to match the largest one, we duplicate the spectrogram and append it to itself, as the sizes of spectrograms may differ.
        \item We utilise ResNet-34 based pre-trained model to extract $300\times257$ dimensional spectrogram.
    \end{itemize}

\begin{table}[!h]

\centering

\centering
footnotesize
\begin{tabular}{|c|ccc|ccc|}
\hline
 \textbf{Dataset}& \multicolumn{3}{c|}{\textbf{FakeAvCeleb}}                            & \multicolumn{3}{c|}{\textbf{SWAN-DF}}                            \\ \hline
 Split& \multicolumn{1}{c|}{\textbf{Train}} & \multicolumn{1}{c|}{\textbf{Val}} & \textbf{Test}  & \multicolumn{1}{c|}{\textbf{Train}} & \multicolumn{1}{c|}{\textbf{Val}} & \textbf{Test}  \\ \hline
Real & \multicolumn{1}{c|}{13320} & \multicolumn{1}{c|}{1480} & 100 & \multicolumn{1}{c|}{11760} & \multicolumn{1}{c|}{3920} & 560 \\ \hline
 Fake& \multicolumn{1}{c|}{15327} & \multicolumn{1}{c|}{1670} & 4047 & \multicolumn{1}{c|}{10335} & \multicolumn{1}{c|}{3648} & 4256 \\ \hline
\end{tabular}
\caption{Dataset distribution after applying augmentation}
\label{tab:data_aug}

\centering
\begin{tabular}{|c|c|c|c|}
\hline
\textbf{Split} &\textbf{Train}  &\textbf{Val}  &\textbf{Test}  \\ \hline
Real &25080  &5400  &660  \\ \hline
 Fake&25662  &5318  &8303  \\ \hline
Total &50742  &10718  &8963  \\ \hline
\end{tabular}
\caption{Total video count after augmentation}
\label{tab:total_aug_final}

\end{table}

\section{Variance and MSE analysis of STGS and GRMC estimators}

\begin{figure*}[!h]
  \centering
   \includegraphics[width=1\linewidth]{images/variance_analysis.png}

   \caption{Variance analysis for two learnable parameters $\alpha$ and $\gamma$ for both GRMC and STGS estimators}
   \label{fig:variance_analysis}
\end{figure*}

\begin{figure*}[!h]
  \centering
   \includegraphics[width=1\linewidth]{images/mse_analysis.png}

   \caption{MSE analysis for STGS and GRMC estimators}
   \label{fig:mse_analysis}
\end{figure*}

Figures \ref{fig:variance_analysis} and \ref{fig:mse_analysis} present a comparative analysis of variance and mean squared error (MSE) for STGS and GRMC estimators. These results support propositions 1 and 2, demonstrating the reduced variance and MSE achieved by the GRMC estimator compared to STGS.

    
    

\begin{figure*}[!h]
    \centering
    \begin{subfigure}[b]{0.3\textwidth}
        \centering
        \includegraphics[width=1\textwidth]{images/ElasticTransformation.jpg}
        \caption*{(a)}
    \end{subfigure}
    \hfill
    \begin{subfigure}[b]{0.3\textwidth}
        \centering
        \includegraphics[width=\textwidth]{images/Add.jpg}
        \caption*{(b)}
    \end{subfigure}
    \hfill
    \begin{subfigure}[b]{0.3\textwidth}
        \centering
        \includegraphics[width=\textwidth]{images/CornerCrop_bottom.jpg}
        \caption*{(c)}
    \end{subfigure}
    
    \begin{subfigure}[b]{0.3\textwidth}
        \centering
        \includegraphics[width=\textwidth]{images/GaussianBlur.jpg}
        \caption*{(d)}
    \end{subfigure}
    \hfill
    \begin{subfigure}[b]{0.3\textwidth}
        \centering
        \includegraphics[width=\textwidth]{images/CornerCrop.jpg}
        \caption*{(e)}
    \end{subfigure}
    \hfill
    \begin{subfigure}[b]{0.3\textwidth}
        \centering
        \includegraphics[width=\textwidth]{images/PiecewiseAffineTransform.jpg}
        \caption*{(f)}
    \end{subfigure}
    
    \caption{Sample images obtained after applying different augmentation techniques: (a) Elastic transformation (b) Add (c) Corner Crop (d) Gaussian blur (e) Center crop (f) Piecewise affine transformation }
    \label{fig:eleven_images}
\end{figure*}

\section{Architectures search}
\subsection{Architectures search on cross dataset generalization}
Figures \ref{fig:arch_fakeav} and \ref{fig:arch_swan} illustrate the generated architectures when trained exclusively on the FakeAVCeleb and SWAN-DF datasets, respectively.
\begin{figure}[htp]
\centering
\begin{minipage}{0.45\textwidth}
\includegraphics[width=\textwidth]{images/arch_fakeav.pdf}
\caption{Architecture obtained when trained only on FakeAVCeleb database}
\label{fig:arch_fakeav}
\end{minipage}\hfill
\begin{minipage}{0.45\textwidth}
\includegraphics[width=\textwidth]{images/arch_swandf.pdf}
\caption{Architecture obtained when trained only on SWAN-DF database}
\label{fig:arch_swan}
\end{minipage}\par
\end{figure}

\subsection{Architectures obtained for different temperature and Monte carlo samples}
Figures \ref{fig:arch_temp0.1}, \ref{fig:arch_temp0.5}, and \ref{fig:arch_temp1} display various architectures generated through the ablation study using different temperature and Monte carlo sampling parameters.

\begin{figure}[htp]
\centering
\begin{minipage}{0.45\textwidth}
\includegraphics[width=\textwidth]{images/temp0.1_1000.pdf}
\caption{Architecture obtained when $\lambda=0.1$ and K=1000}
\label{fig:arch_temp0.1}
\end{minipage}\hfill
\begin{minipage}{0.45\textwidth}
\includegraphics[width=\textwidth]{images/temp_0.5_10.pdf}
\caption{Architecture obtained when $\lambda=0.5$ and K=10}
\label{fig:arch_temp0.5}
\end{minipage}\par
\vskip\floatsep
\includegraphics[width=0.45\textwidth]{images/temp_1.0_100.pdf}
\caption{Architecture obtained when $\lambda=1.0$ and K=100}
\label{fig:arch_temp1}
\end{figure}

\subsection{ROC curves for real and fake data}
Figures \ref{fig:roc_real} and \ref{fig:roc_fake} illustrate the ROC curves for Softmax, STGS-BMNAS, and GRMC-BMNAS. The GRMC-BMNAS consistently outperforms the other methods in terms of roc for both real and fake classes.

\begin{figure}[htp]
\centering
\begin{minipage}{0.45\textwidth}
\includegraphics[width=\textwidth]{images/auc_real.png}
\caption{ROC curve comparison for the real class: Softmax, STGS-BMNAS, and GRMC-BMNAS.}
\label{fig:roc_real}
\end{minipage}\hfill
\begin{minipage}{0.45\textwidth}
\includegraphics[width=\textwidth]{images/auc_fake.png}
\caption{ROC curve comparison for the fake class: Softmax, STGS-BMNAS, and GRMC-BMNAS}
\label{fig:roc_fake}
\end{minipage}\par
\end{figure}

\subsection{Stable Variance and Improved Performance of GRMC Estimator at Low Temperatures}
\begin{itemize}
    \item \textbf{Improvement below t=0.1:} The trend from the ablation study indicates that the performance of the GRMC estimator improves as the temperature $\lambda$ decreases below 0.1.
    \item \textbf{Variance Behavior:}For the Gumbel Softmax (GS) and Straight-Through Gumbel Softmax (GS-ST) estimators, the variance grows unbounded as the temperature 
t decreases, following an asymptote of $\mathcal{O}(\frac{1}{\lambda})$ approaches zero, the variance increases significantly.
\item In contrast, for the GRMC estimator, the variance does not grow unbounded as $\lambda$ decreases. This suggests that the GRMC estimator has a more stable variance behavior at low temperatures, making it potentially more robust in such conditions.
\end{itemize}

\subsection{Failure cases}
\begin{figure*}[!h]
  \centering
   \includegraphics[width=1\linewidth]{images/face_images.jpg}

   \caption{Failure cases for the proposed approach. All the videos comes from FakeAVCeleb database}
   \label{fig:falure_cases}
\end{figure*}

While no failure cases were observed in the frontal video dataset SWAN-DF, some failures were identified in the FakeAVCeleb database. These instances are highlighted in Figure \ref{fig:falure_cases}.

\section{Limitations}
Our current method employs a manual tuning of temperature and Monte Carlo samples to balance exploration and exploitation. To enhance this, we propose adaptive strategies. First, we advocate for dynamically adjusting the number of Monte Carlo samples based on the estimated variance of the current solutions, prioritizing efficiency. Second, we suggest replacing the static temperature with a tempered Gumbel softmax approach, allowing for a more nuanced exploration-exploitation balance. This adaptive approach is expected to improve the search efficiency and solution quality.